\documentclass[letterpaper,10 pt, conference]{ieeeconf}
\usepackage{graphicx}
\usepackage{amsmath}
\usepackage{url}
\usepackage{multirow}
\IEEEoverridecommandlockouts

\begin{document}
\title{\LARGE \bf Learning Front-end Filter-bank Parameters using Convolutional Neural Networks for Abnormal Heart Sound Detection}

\author{Ahmed Imtiaz Humayun$^{1}$, Shabnam Ghaffarzadegan$^{2}$, Zhe Feng$^{2}$ and Taufiq Hasan$^{1}$\thanks{$^{1}$Taufiq Hasan and Ahmed Imtiaz Humayun are affiliated with mHealth Laboratory, Department of Biomedical Engineering, Bangladesh University of Engineering and Technology (BUET), Dhaka - 1205, Bangladesh. Email: {\tt\scriptsize taufiq@bme.buet.ac.bd, imtiaz@mhealth.buet.ac.bd}}%
\thanks{$^{2}$Shabnam Ghaffarzadegan and Zhe Feng are with the Human Machine Interaction Group-2, Robert Bosch Research and Technology Center (RTC), Palo Alto, CA - 94304, USA. Email: {\tt\scriptsize \{shabnam.ghaffarzadegan,zhe.feng2\}@us.bosch.com}}%
}
\maketitle
\pubid{\makebox[\linewidth]{
\copyright~2018
IEEE. Accepted for publication in the 40th International Engineering in Medicine and Biology Conference (EMBC).\hfill} }


\begin{abstract}
\pubidadjcol
Automatic heart sound abnormality detection can play a vital role in the early diagnosis of heart diseases, particularly in low-resource settings. 
The state-of-the-art 
algorithms for this task utilize a set of Finite Impulse Response (FIR) band-pass filters as a front-end followed by a Convolutional Neural Network (CNN) model. 
In this work, we propound a novel CNN architecture that integrates the front-end band-pass filters within the network using time-convolution (tConv) layers, which enables the FIR filter-bank parameters to become learnable. 
Different initialization strategies for the learnable filters, including random parameters and a set of predefined FIR filter-bank coefficients, are examined. Using the proposed tConv layers, we add constraints to the learnable FIR filters to ensure linear and zero phase responses. Experimental evaluations are performed on a balanced 4-fold cross-validation task prepared using the PhysioNet/CinC 2016 dataset. Results demonstrate that the proposed models yield superior performance compared to the state-of-the-art system, while the linear phase FIR filter-bank method provides an absolute improvement of 9.54\% over the baseline in terms of an overall accuracy metric.

\end{abstract}


\IEEEpeerreviewmaketitle

\section{Introduction}
Cardiovascular diseases (CVDs) are responsible for about 17.7 million deaths every year, representing 31\% of the global mortality \cite{whofact}. Cardiac auscultation is the most popular non-invasive and cost-effective procedure for the early diagnosis of various heart diseases. However, effective cardiac auscultation requires trained physicians, a resource which is limited especially in low-income countries of the world \cite{alam2010cardiac}. 
Thus, machine learning based automated heart sound classification systems implemented with a smart-phone attachable digital stethoscope in the point-of-care locations can be of significant impact for early diagnosis of cardiac diseases, particularly for countries that suffer from a shortage and geographic mal-distribution of skilled physicians.

Automated classification of the Phonocardiogram (PCG), i.e., the heart sound, have been extensively studied and researched in the past few decades. Analysis of the PCG can be broadly divided into two principal areas: (i) segmentation of the PCG signal, i.e., detection of 
the first and second heart sounds (S1 and S2), and (ii) classification of recordings as pathologic or physiologic. Conventionally heart sound classification methods employed Artificial Neural Networks (ANN) \cite{uuguz2012ANN}, Support Vector Machines (SVM) \cite{gharehbaghi2015SVM} and Hidden Markov Models (HMM) \cite{saraccouglu2012HMM}. The 2016 Physionet/CinC Challenge released an archive of $4430$ PCG recordings, which is the most extensive open-source heart sound dataset to date.
Time, frequency and statistical features \cite{homsi2017}, Mel-frequency Cepstral Coefficients (MFCC) \cite{bobillo2016}, and Continuous Wavelet Transform (CWT), were some of the commonly used features by the PhysioNet challenge entrants. 
Among the top scoring systems, Maknickas et al. \cite{maknickas2017} extracted Mel-frequency Spectral Coefficients (MFSC) from unsegmented signals and used a 2D CNN. 
Plesinger et al. \cite{plesinger2017} proposed a novel segmentation method, a histogram based feature selection method and parameterized sigmoid functions per feature, to discriminate between classes. 
Various machine learning algorithms including SVM \cite{whitaker2017}, k-Nearest Neighbor (k-NN) \cite{bobillo2016}, Multilayer Perceptron (MLP) \cite{kay2017,zabihi2016}, Random Forest \cite{homsi2017}, 1D \cite{potes2016ensemble} and 2D CNNs \cite{maknickas2017}, and Recurrent Neural Network (RNN) \cite{yang2016classification} were employed in the challenge. A good number of submissions used an ensemble of classifiers with a voting algorithm \cite{homsi2017,kay2017,zabihi2016,potes2016ensemble}. The best performing system was presented by Potes et al. \cite{potes2016ensemble} 
that combined a 1D-CNN model with an Adaboost-Abstain classifier using a threshold based voting algorithm.
\pubidadjcol
Filter-banks are used as a standard pre-processing step during audio feature engineering and are also incorporated in \cite{potes2016ensemble} before the 1D-CNN. However, no particular physiological significance of the filter-bank structure and their cutoff frequency definitions were presented. In this work, we propose a CNN based Finite Impulse Response (FIR) filter-bank front-end, that learns the frequency characteristics of the FIR filters, such that they are more effective in distinguishing pathologic heart sounds.

\section{Dataset}
\subsection{PhysioNet/CinC Challenge Dataset}
The 2016 PhysioNet/CinC Challenge dataset \cite{liu2016open} is an accumulation of PCG recordings from seven different research groups, consisting of an open training set and a hidden test set. The dataset contains six subsets (a-f) corresponding to the contributing groups. The training data contains $3153$ heart sound recordings collected from $764$ patients with a total number of $84,425$ cardiac cycles ranging from $35$ to $159$ bpm. The dataset is class unbalanced with $2488$ \emph{Normal} and $665$ \emph{Abnormal} heart sound recordings. 
\subsection{Dataset Preparation for Cross-validation}\label{inhousedata}
Considering the fact that the dataset is unbalanced and the number of recordings is small, we divided the dataset into 4 folds for cross-validation, with balanced validation sets (equal number of normal and abnormal recordings). 
A validation set of $301$ recordings was already provided by PhysioNet (Fold 0). The rest of the three folds are created by random sampling without replacement. 
\begin{figure}[t]
\includegraphics[width=\linewidth,trim={0 1.5cm 0 0}]{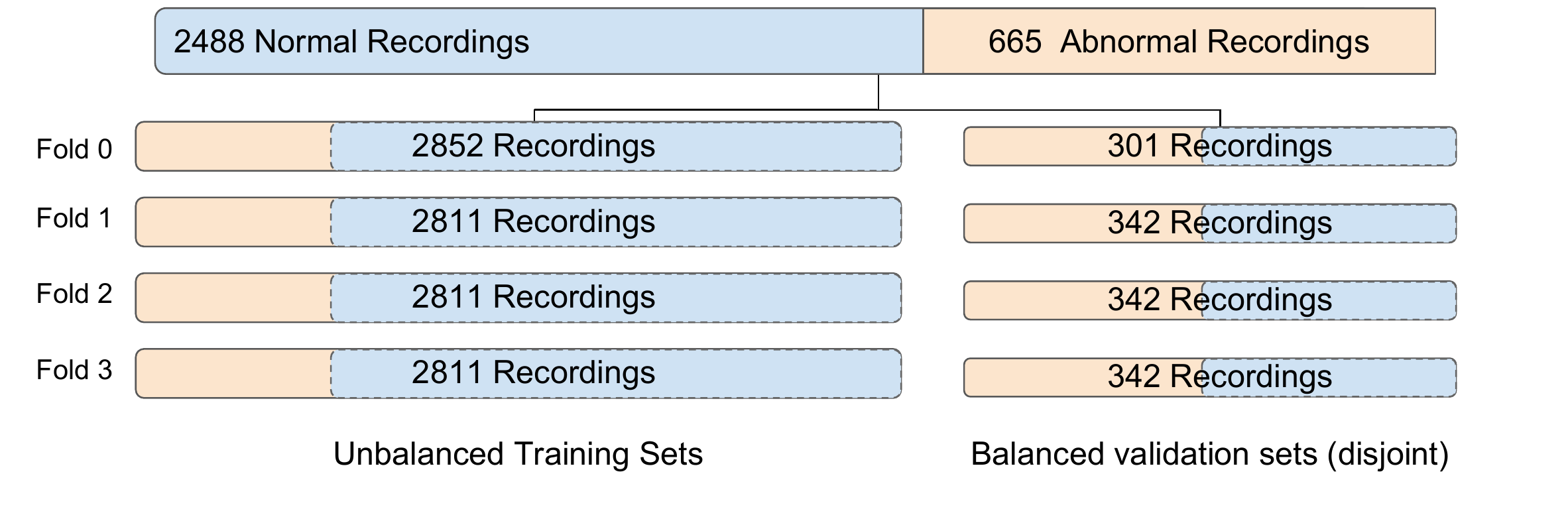}
\centering
\caption{The PCG recordings in the prepared 4-fold cross validation dataset.}
\label{foldsplit}
\vspace{-3mm}
\end{figure}

\section{Proposed Method}


\subsection{Baseline Implementation}\label{baseline}
Our baseline system follows the CNN system proposed in the top scoring solution \cite{potes2016ensemble} of the Physionet/CinC 2016 challenge.
First we pre-process the signal, to decompose it into into four frequency bands ($25-45$, $45-80$, $80-200$, $200-500$ Hz).
Next, cardiac cycles are extracted using PCG segmentation \cite{springer2016segmentation} and zero padded to be $2.5$s in length.
Four different bands of each cardiac cycle are fed into $4$ different input branches of the 1D-CNN. Each branch has two convolutional layers of kernel size $5$, followed by a Rectified Linear Unit (ReLU) activation and a max-pooling of $2$. The first convolutional layer has $8$ filters while the second has $4$. The outputs of the four branches are fed to an MLP network after being flattened and concatenated. The MLP network has a hidden layer of $20$ neurons with ReLU activation and a single neuron as output with sigmoid activation. 
Adam optimization is used with binary cross-entropy as the loss function. The resulting model provides predictions on every cardiac cycle, which are averaged over the entire recording and rounded for inference.


\subsection{Learnable Filter-banks: The tConv Layer}\label{convFIR}
\begin{figure}[t]
\includegraphics[width=7cm,keepaspectratio,trim={0 1cm 0 0}]{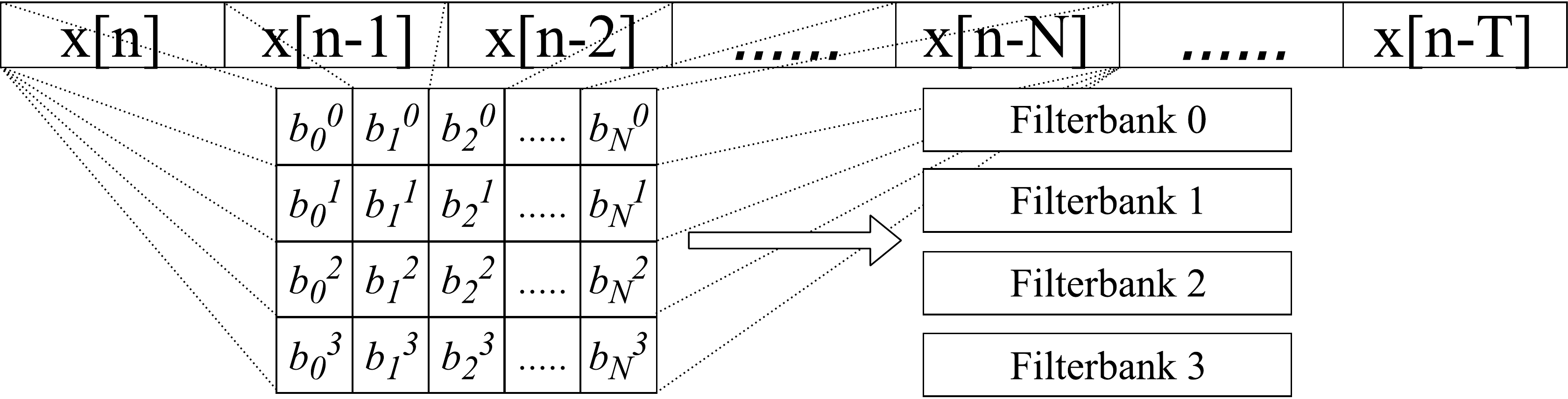}
\centering
\caption{Operation of a tConv layer as an FIR filter-bank.}
\label{tconv}
\vspace{-3mm}
\end{figure}

\begin{figure*}[t]
\includegraphics[width=\textwidth,trim={0 .5cm 0 0}]{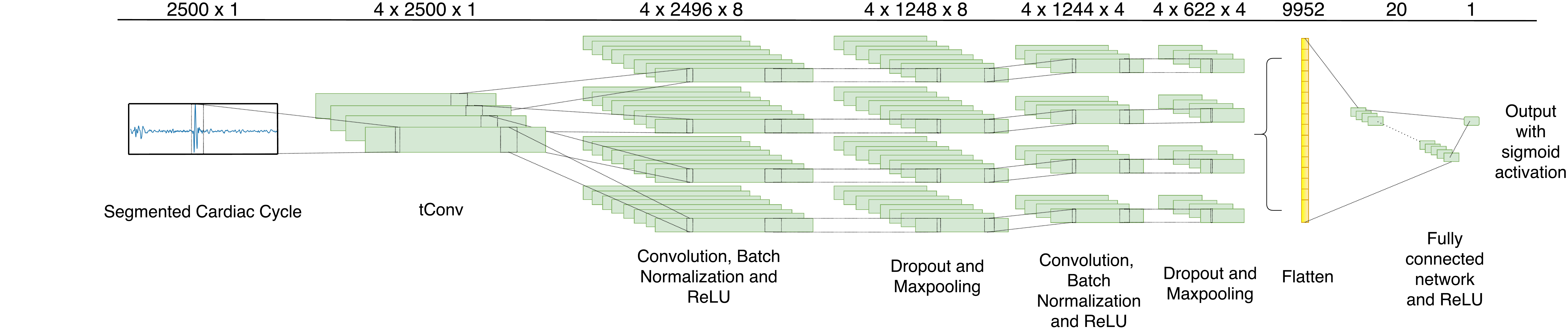}
\centering
\caption{Proposed CNN Model Architecture including a Learnable Front-end Filter-bank (tConv Layers).}
\label{propmod}
\end{figure*}

For a causal discrete-time FIR filter of order $N$ with filter coefficients $b_{0}, b_{1}, \dots b_{N}$, the output samples $y[n]$ are obtained by a weighted sum of the most recent samples of the input signal $x[n]$. This can be expressed as: 
\begin{eqnarray}
y[n] &=& b_{0}x[n]+b_{1}x[n-1]+.....+b_{N}x[n-N] \nonumber\\
 &=& \sum_{i=0}^{N}b_{i}x[n-i].
\end{eqnarray}
Through a local connectivity pattern of neurons between adjacent layers, a 1D-CNN performs cross-correlation between it's input and it's kernel. The output of a convolutional layer, with a kernel of odd length $N+1$, can be expressed as: 
\begin{align}\label{conveqn}
\small
y[n] & = b_{0}x[n+\tfrac{N}{2}]+b_{1}x[n+\tfrac{N}{2}-1]+....+b_{\tfrac{N}{2}}x[n]+.... \nonumber\\
& +b_{N-1}x[n-\tfrac{N}{2}+1]+b_{N}x[n-\tfrac{N}{2}] \nonumber\\
& = \sum_{i=0}^{N}b_{i}\hspace{0.5mm}x[n+\tfrac{N}{2}-i]
\end{align}
where $b_{0}, b_{1}, ... b_{N}$ are the kernel weights. Considering a causal system the output of the convolutional layer becomes:
\begin{equation}
y[n - \tfrac{N}{2}] = \sigma\left(\beta +\sum_{i=0}^{N}b_{i}x[n-i]\right)
\end{equation}
where $\sigma(\cdot)$ is the activation function and $\beta$ is the bias term. Therefore, a 1D convolutional layer with linear activation and zero bias, acts as an FIR filter with an added delay of $N/2$ \cite{matei2006CNNFIR}. We denote such layers as time-convolutional (tConv) layers (Fig. \ref{tconv}) \cite{sainath2015google}. Naturally, the kernels of these layers (similar to filter-bank coefficients) can be updated with Stochastic Gradient Descent (SGD). This enables the tConv layers to learn coefficients that decompose the signal into pathologically meaningful sub-bands. The front-end filter-banks used in \cite{potes2016ensemble} were coalesced with the model architecture using tConv layers. The resulting architecture is shown in Fig. \ref{propmod}.

During implementation of the tConv network, further modifications over the baseline model was made that include: addition of the initialization scheme presented by He et al. \cite{he2016deep}, dropping out activations before max-pooling, and addition of batch-normalization after every convolutional layer. 
The hyper-parameters are re-tuned for optimal performance as shown in Table \ref{hyparam}, 
using Tree of Parzen Estimators \cite{bergstra2011NIPS}.

\begin{table}[b]
\centering
\caption{Hyperparameters of the 1D-CNN Model}
\label{hyparam}
\begin{tabular}{|r|l|}
\hline
Parameter					& Value \\
\hline
Learning  Rate              & 0.0012843784 \\ 
Learning Rate Decay         & 0.0001132885 \\ 
Dropout after Conv Layers & 50\%         \\ 
L2-regularization in Conv layers      & 0.0486         \\ 
Pool size                   & 2            \\ \hline
\end{tabular}
\end{table}

\begin{table}[ht]
\centering
\caption{Experimental Results on the 4-Fold Cross-Validation Data}
\label{metrics}
\resizebox{\linewidth}{!}{%
\begin{tabular}{|c|c|c|c|c|c|c|c|}
\hline
Model & Fold & \begin{tabular}[c]{@{}c@{}}Sensitivity\\ (\%)\end{tabular} & \begin{tabular}[c]{@{}c@{}}Specificity\\ (\%)\end{tabular} & \begin{tabular}[c]{@{}c@{}}Macc\\ (\%)\end{tabular} & \begin{tabular}[c]{@{}c@{}}Cross-fold\\ Sesitivity\\ (\%)\end{tabular} & \begin{tabular}[c]{@{}c@{}}Cross-fold\\ Specificity\\ (\%)\end{tabular} & \begin{tabular}[c]{@{}c@{}}Cross-fold\\ Macc\\ (\%)\end{tabular} \\ \hline
\multirow{4}{*}{Baseline \cite{potes2016ensemble}} & 0 & 63.76 & 81.11 & 72.44 & \multirow{4}{*}{\begin{tabular}[c]{@{}c@{}}64.06\\ ($\pm$2.5)\end{tabular}} & \multirow{4}{*}{\begin{tabular}[c]{@{}c@{}}91.06\\ ($\pm$6.8)\end{tabular}} & \multirow{4}{*}{\begin{tabular}[c]{@{}c@{}}77.56\\ ($\pm$3.4)\end{tabular}} \\ 
 & 1 & 64.07 & 94.15 & 79.11 &  &  &  \\ 
 & 2 & 61.06 & 96.57 & 78.82 &  &  &  \\ 
 & 3 & 67.33 & 92.39 & 79.86 &  &  &  \\ \hline
\multirow{4}{*}{\begin{tabular}[c]{@{}c@{}}tConv\\ Non-Learn\end{tabular}} & 0 & 89.30 & 56.26 & 72.78 & \multirow{4}{*}{\begin{tabular}[c]{@{}c@{}}88.74\\ ($\pm$1.53)\end{tabular}} & \multirow{4}{*}{\begin{tabular}[c]{@{}c@{}}79.94\\ ($\pm$16)\end{tabular}} & \multirow{4}{*}{\begin{tabular}[c]{@{}c@{}}84.34\\ ($\pm$7.85)\end{tabular}} \\ 
 & 1 & 89.39 & 86.54 & 87.97 &  &  &  \\ 
 & 2 & 89.80 & 90.48 & 90.14 &  &  &  \\ 
 & 3 & 86.47 & 86.47 & 86.47 &  &  &  \\ \hline
\multirow{4}{*}{\begin{tabular}[c]{@{}c@{}}tConv-\\ FIR Init\end{tabular}} & 0 & 91.57 & 57.14 & 74.36 & \multirow{4}{*}{\begin{tabular}[c]{@{}c@{}}87.72\\ ($\pm$2.9)\end{tabular}} & \multirow{4}{*}{\begin{tabular}[c]{@{}c@{}}83.39\\ ($\pm$17.5)\end{tabular}} & \multirow{4}{*}{\begin{tabular}[c]{@{}c@{}}85.55\\ ($\pm$7.5)\end{tabular}} \\ 
 & 1 & 86.29 & 91.31 & 88.81 &  &  &  \\ 
 & 2 & 88.14 & 93.15 & 90.64 &  &  &  \\ 
 & 3 & 84.87 & 91.98 & 88.42 &  &  &  \\ \hline
\multirow{4}{*}{\begin{tabular}[c]{@{}c@{}}LP-tConv-\\ FIR Init\end{tabular}} & 0 & 88.45 & 65.65 & 77.05 & \multirow{4}{*}{\begin{tabular}[c]{@{}c@{}}90.91\\ ($\pm$2.4)\end{tabular}} & \multirow{4}{*}{\begin{tabular}[c]{@{}c@{}}83.29\\ ($\pm$11.8)\end{tabular}} & \multirow{4}{*}{\begin{tabular}[c]{@{}c@{}}87.10\\ ($\pm$6.79)\end{tabular}} \\ 
 & 1 & 93.81 & 88.38 & 91.10 &  &  &  \\ 
 & 2 & 91.72 & 90.97 & 91.35 &  &  &  \\ 
 & 3 & 89.64 & 88.14 & 88.89 &  &  &  \\ \hline
\multirow{4}{*}{\begin{tabular}[c]{@{}c@{}}ZP-tConv-\\ FIR Init\end{tabular}} & 0 & 90.73 & 56.65 & 73.69 & \multirow{4}{*}{\begin{tabular}[c]{@{}c@{}}89.52\\ ($\pm$1.1)\end{tabular}} & \multirow{4}{*}{\begin{tabular}[c]{@{}c@{}}81.41\\ ($\pm$16.6)\end{tabular}} & \multirow{4}{*}{\begin{tabular}[c]{@{}c@{}}85.47\\ ($\pm$8)\end{tabular}} \\ 
 & 1 & 89.22 & 90.31 & 89.77 &  &  &  \\ 
 & 2 & 89.97 & 91.81 & 90.89 &  &  &  \\ 
 & 3 & 88.14 & 86.88 & 87.51 &  &  &  \\ \hline
\multirow{4}{*}{\begin{tabular}[c]{@{}c@{}}LP-tConv-\\ Rand Init\end{tabular}} & 0 & 73.23 & 79.84 & 76.53 & \multirow{4}{*}{\begin{tabular}[c]{@{}c@{}}84.01\\ ($\pm$8)\end{tabular}} & \multirow{4}{*}{\begin{tabular}[c]{@{}c@{}}86.79\\ ($\pm$5.8)\end{tabular}} & \multirow{4}{*}{\begin{tabular}[c]{@{}c@{}}85.40\\ ($\pm$6.2)\end{tabular}} \\ 
 & 1 & 92.40 & 86.13 & 89.26 &  &  &  \\ 
 & 2 & 86.13 & 93.98 & 90.06 &  &  &  \\ 
 & 3 & 84.29 & 87.22 & 85.76 &  &  &  \\ \hline
\end{tabular}%
}
\end{table}

\begin{figure}[b]
\includegraphics[width=\linewidth,keepaspectratio]{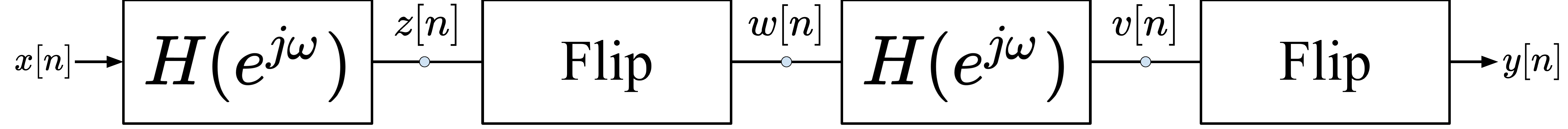}
\centering
\caption{Forward reverse filtering in a Zero Phase tConv (ZP-tConv) layer.}
\label{filtfilt}
\end{figure}

\subsection{Variants of the tConv Layer}
\subsubsection{Linear Phase tConv}\label{LPtConv}
The FIR intuition of tConv layers institutes new insights into the frequency and phase response of the kernel. Especially, large kernels can introduce significant phase distortion into their activations. The phase response of a filter indicates the phase shift in radians that each input component sinusoid will undergo. A convolutional kernel with non-linear phase would introduce a temporal shift between the high frequency (e.g., murmurs) and low frequency (e.g.,  systole and diastole) patterns in the PCG signal. To mitigate the effect, we propose a novel convolutional architecture termed linear phase tConv. Linear Phase (LP) is the condition when the phase response of a filter is a linear function of frequency (excluding phase wraps at +/- $\pi$ radians). A kernel with symmetric weights around its center would have linear phase, i.e., it would introduce an equal delay for all of the passing frequencies/patterns, ensuring no distortion. Results are further discussed in Sec. \ref{kernelinit}.

\subsubsection{Zero Phase tConv}
Zero phase (ZP) filter is a special case of a linear phase FIR filter, where the phase response is nullified. Incorporating a forward-reverse convolution into tConv layers \cite{shi2006ZPtconv}, we propose a zero phase tConv layer, the operation of which is shown in Fig \ref{filtfilt}. If $x[n]$ is the input signal, $h[n]$ is the impulse response of the kernel, and $X(e^{j\omega})$ and $Y(e^{j\omega})$ are fourier transforms of $x[n]$ and $h[n]$:
\begin{equation}
\begin{split}
Y(e^{j\omega})&=X(e^{j\omega}).H^{*}(e^{j\omega}).H(e^{j\omega})\\
&=X(e^{j\omega})|H(e^{j\omega})|^{2}\\
\text{where}  \quad & H(e^{j\omega})=|H(e^{j\omega})|\angle H(e^{j\omega})
\end{split}
\end{equation}
Note that, the flip operation in time domain is equivalent to taking the complex conjugate in the frequency domain. 
Therefore, the effect of a ZP-tConv is just a multiplication by the squared magnitude in the frequency domain.

\section{Results and Discussion}
\subsection{Experimental Evaluation}
The performance of the proposed methods are evaluated and compared with the baseline on our 4-fold cross validation dataset (Sec. \ref{inhousedata}). The loss function was weighted during training to draw emphasis on abnormal recordings as they represented only 21\% of the data. As performance metrics, sensitivity, specificity and Macc (mean of sensitivity and specificity) are calculated and averaged over the 4-folds. The proposed tConv model is also evaluated with the FIR filter parameters fixed as described in Sec. \ref{baseline} (tConv Non-Learn). The results are summarized in Table \ref{metrics}. 

From the results, we observe that the best system attained an averaged cross-fold Macc of 87.10{\small($\pm6.79$)}\% using the proposed LP-tConv approach with FIR initialization. This represents an absolute improvement of 9.54\% over the baseline CNN system \cite{potes2016ensemble}. Other variants of the proposed tConv systems also provided superior performance compared to the baseline as seen in Table \ref{metrics}.



\begin{figure}[b]
\includegraphics[width=\linewidth]{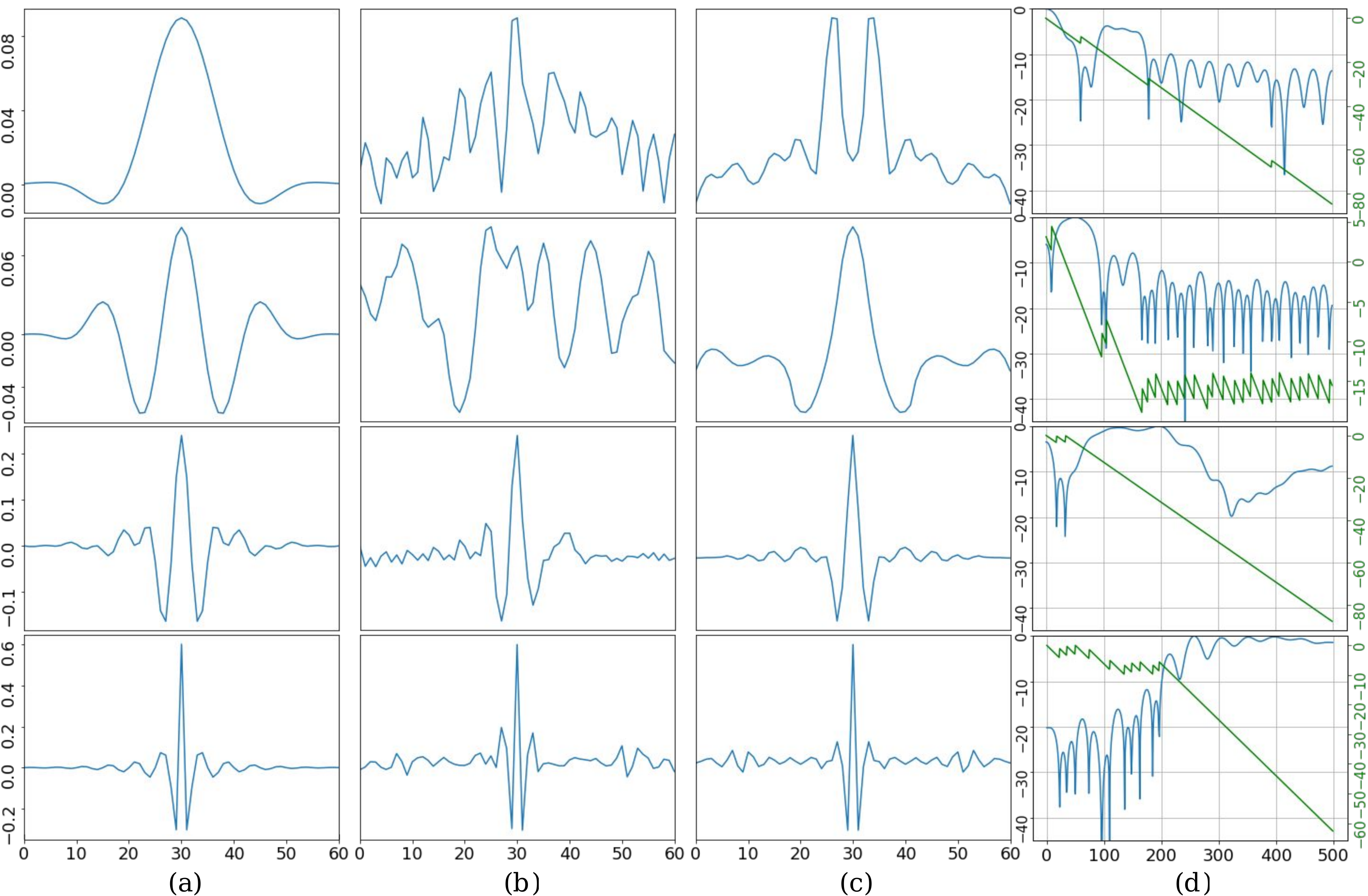}
\centering
\caption{In the panel of figures, each row represent input branches (1-4) of the CNN model. For each input branch, the columns represent: (a) Initial FIR coefficients, (b) Learned FIR coefficients in tConv, (c) Learned FIR coefficients in LP-tConv (d) Magnitude (blue) and phase response (green) of the learned filters via LP-tConv.}
\label{diffinit}
\end{figure}

\subsection{Kernel Initialization for tConv layers}\label{kernelinit}

As discussed in \ref{LPtConv} making the kernel symmetric reduces phase distortion which has an additional benefit of requiring half the number of learnable parameters in a tConv layer. Compared to ZP-tConv, learning symmetric pattern improves the Macc metric (Table \ref{metrics}). We also experimented with zero, random and FIR initialization (initialized with FIR coefficients as in Sec. \ref{baseline}) schemes. Visualizing the learned coefficients and their frequency responses (Fig \ref{diffinit}), we observe that higher frequency coefficients are less affected by training compared to lower frequency coefficients.



\subsection{Dataset Variability and Result Analysis}
\begin{figure}[t]
\includegraphics[width=\linewidth,keepaspectratio]{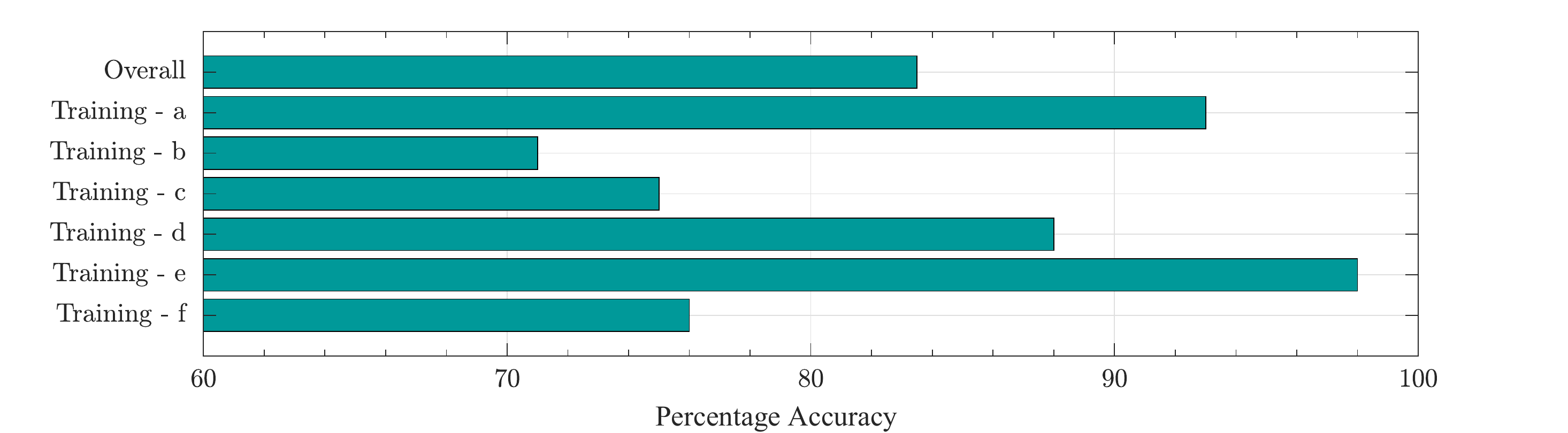}
\centering
\caption{Validation Accuracy per cardiac cycle for LP-tConv on different training subsets of the Physionet Heart Sound Dataset.}
\label{chart}
\end{figure}

\begin{figure}[t]
\includegraphics[width=0.95\linewidth,trim={1.5cm 13cm 2cm 1.5cm},clip]{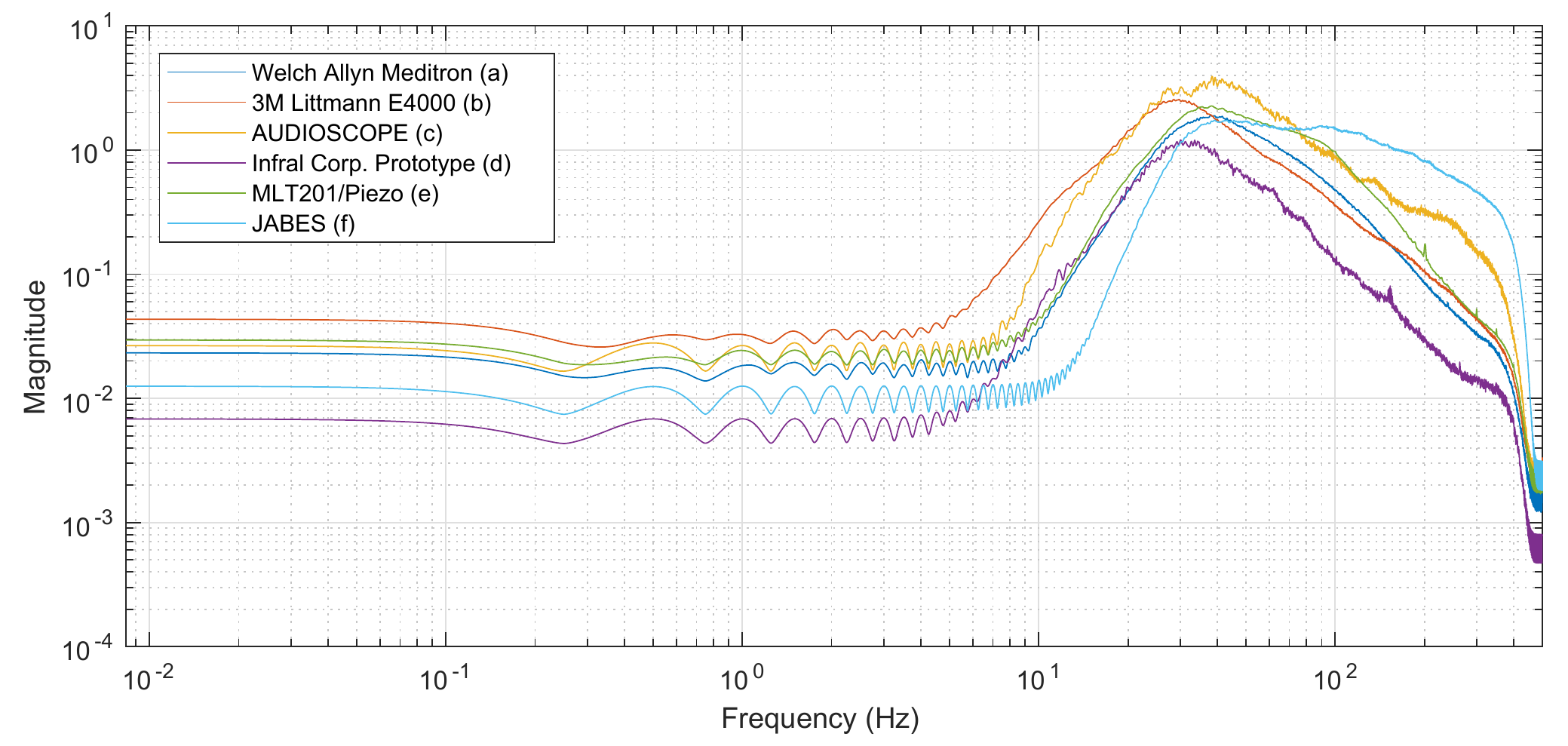}
\vspace{-4mm}
\centering
\caption{Long-term Spectral Average (LTSA) of normal heart sound recordings captured using different sensors.}
\label{ltsa}
\vspace{-3mm}
\end{figure}

In Fig. \ref{chart}, we compare the performance of the proposed LP-tConv system over different data subsets. The model performed lowest on the SUAHSDB (training-f) subset of the PhysioNet data. Performance on fold 0 is substandard compared to the other folds (Table \ref{metrics}). We were unable to find any correlation between signal quality and the model performance. A Long-Term Spectral Average (LTSA) \cite{byrne1994LTSA} over normal heart sound PCG showed differences in the frequency characteristics of sensors used during recording as seen in Fig. \ref{ltsa}. Here, a distinct difference is visible between the frequency envelope of JABES electronic stethoscope and the other stethoscopes. Approximately 67\% of the training data belongs to training-e, which created a dependency of the model towards the characteristics of this subset. Besides, training-e was recorded using a unique piezoelectric sensor based stethoscope \cite{liu2016open}, which may also be contributing towards the suboptimal generalization. Fold 0 contained a lower percentage of training-e in its validation set, which explains the poor validation performance.

 

\section{Conclusion}
In this study, we have proposed novel tConv layers with CNN as learnable filter-banks for normal-abnormal heart sound classification. 
Different initialization strategies have been examined for the tConv layers while constraints have been added to ensure a zero and linear phase response in the resulting FIR filters.
Experimental results using the proposed architecture shows significant improvements compared to state-of-the-art solutions with respect to various performance metrics on a cross-validation task prepared using the PhysioNet heart sound challenge dataset.



\bibliographystyle{IEEEtran}
\bibliography{refs.bib}
\end{document}